\documentclass[letterpaper, 10 pt, conference]{ieeeconf}  

\IEEEoverridecommandlockouts             

\usepackage{comment}
\usepackage{amsmath}
\usepackage{amssymb}
\usepackage{graphicx}
\usepackage{url}
\usepackage[hidelinks]{hyperref}
\usepackage{color}
\usepackage{booktabs}
\usepackage{tikz}
\usetikzlibrary{shapes,arrows,positioning,calc}
\usepackage{siunitx}
\usepackage[export]{adjustbox}
\usepackage{enumerate}
\usepackage{algorithm2e}
\usepackage{nicefrac}
\usepackage{calc}
\usepackage{leftidx}

\newcommand{\Be}{\begin{equation}}
\newcommand{\Ee}{\end{equation}}
\newcommand{\bs}[1]{\boldsymbol{#1}}

\newcommand{\T}[2]{\leftidx{_\text{#1}}{\boldsymbol{T}}{^\text{#2}}}
\newcommand{\dT}[2]{\leftidx{_\text{#1}}{\boldsymbol{\xi}}{^\text{#2}}}
\newcommand{\ddT}[2]{\leftidx{_\text{#1}}{\boldsymbol{\dot{\xi}}}{^\text{#2}}}

\title{\LARGE \bf
5G Virtual Reality Manipulator Teleoperation using a Mobile Phone
}

\author{Alexander Werner and William Melek
\thanks{Authors are with the 
        University of Waterloo, Waterloo, Canada.\newline
        {\tt\small name.surname@uwaterloo.ca}}%
}

\begin{document}
\maketitle
\thispagestyle{empty}
\pagestyle{empty}

\begin{abstract}
This paper presents an approach to teleoperate a manipulator using a mobile phone as a leader device. Using its IMU and camera, the phone estimates its Cartesian pose which is then used to to control the Cartesian pose of the robot's tool. The user receives visual feedback in the form of multi-view video - a point cloud rendered in a virtual reality environment. This enables the user to observe the scene from any position. To increase immersion, the robot's estimate of external forces is relayed using the phone's haptic actuator. Leader and follower are connected through wireless networks such as 5G or Wi-Fi. The paper describes the setup and analyzes its performance.
\end{abstract}

\section{INTRODUCTION}
The principal goal of the presented approach is to provide users with a portable and intuitive way of teleoperating a manipulator. With regards to existing teleoperation interfaces, we present an extreme choice. The interface is already in everyone's pocket - a mobile phone. This guarantees portability but requires working within the limitations of the device.

Most manipulator teleoperation products are on the opposite end of the design space: a premium tabletop teleoperation workstation which includes either a virtual reality (\textsl{VR}) headset or an autostereoscopic display for visual feedback combined with a six-degree-of-freedom haptic device equipped with fine-grained position sensing and force feedback abilities. None of these three modalities can be directly replicated on a mobile phone.

\begin{figure}
    \includegraphics[width=\columnwidth]{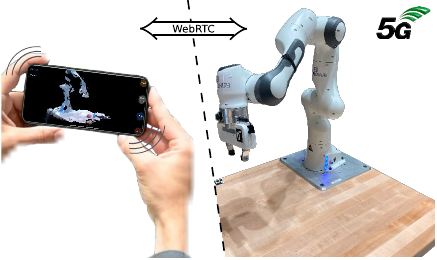}
    \caption{Experimental setup used for evaluation.}
    \label{fig:setup}
\end{figure}

As illustrated in Fig.~\ref{fig:setup}, our approach implements visual feedback by rendering a point cloud of the scene containing the robot and the environment. The user can navigate the scene by moving the phone, which remedies the lack of depth perception on the phone display.
For control, we estimate the pose of the mobile phone using proprioceptive and exteroceptive sensors: the onboard inertial measurement unit (IMU) and camera. With the activation of a virtual clutch, changes in this pose are replicated 1:1 at the position level by the manipulator in both translation
and orientation (SE3) if and as desired.

To relay contact forces to the operator, the built-in haptic actuator is used, but this is hardly equivalent to the force feedback a purpose-designed haptic interface can generate. The actuator is incapable of generating static forces, and only minute dynamic forces in one direction can be generated. We relay static forces with an amplitude modulation scheme driving the haptic actuator. Especially for haptics, a reliable network connection with low latency provided by wireless networks such as 5G is important.

Where can this approach be useful? We think that the best use case is in the temporary control of a manipulator in a scenario where human understanding of the manipulation task can directly be applied. Due to the temporary nature, conventional haptic interfaces are unlikely to be an efficient choice. An example for such a scenario is error recovery for industrial or collaborative manipulators. Our approach allows the operator to intervene remotely using a more intuitive interface than the traditional button-based pendants.

%

The contributions of this paper are the following. To the authors' knowledge, using a pose obtained from visual-inertial fusion on a mobile phone to directly control a manipulator pose in SE3 is novel. This paper also proposes a novel method to encode transient and static contact forces into haptic actuator feedback. Additionally, this paper demonstrates how to stream typical RGB-D point clouds over wireless networks to a mobile device considering bandwidth and computation limits while still achieving real-time rendering.

The remainder of this paper is structured as follows. After a review of the relevant
literature in Sec.~\ref{sec:literature}, Sec.~\ref{sec:design} describes the design choices which led to the approach detailed
in Sec.~\ref{sec:control}. Sec.~\ref{sec:experimental} then describes the experimental setup used
to evaluate the approach generating the data presented in Sec.~\ref{sec:performance}. Finally, 
Sec.~\ref{sec:conclusion} provides a conclusion.

\section{Related Work}
\label{sec:literature}

The authors of \cite{wu2020development} present an approach where only proprioceptive sensors of a mobile device are used to teleoperate
a manipulator. In this approach, the orientation of the mobile device is mapped to the angular velocity
of the manipulator while the translation velocity is controlled via the touch screen. This approach seems quite
intuitive and is well adapted for use as an assistive device for people with limited mobility in the
upper extremities as it does not require to move the device in translation.
Teleoperation of a manipulator using a mobile device also has been presented in \cite{parga2013tele}, the approach uses
proprioceptive sensors only. It maps the translation and orientation offsets of the phone directly to the manipulator.
Both papers focus on control and do not describe any visual or haptic feedback on the mobile device.
Shared autonomy-based teleoperation interfaces and interactive manipulation methods can be aided by object recognition as shown in \cite{teleopcomp2017}.
Other alternative interfaces for teleoperation of a manipulator include using human pose estimation include using RGB-D cameras \cite{handa2020dexpilot,qin2023anyteleop}, or using controllers together with a VR headset \cite{hetrick2020comparing}.

For anchored teleoperation setups, research has provided insights into the requirements of \textsl{haptics} as part of force-feedback \cite{Impact2013}. 
Feedback using haptic actuators has been explored for teleoperation for some time. \cite{Cluttered2017} presents an approach using haptics-enabled armbands and uses hand-tracking for controlling the robot's pose. The paper also shows that having haptic feedback is preferred by the user's teleoperation setup.

\textsl{Visual feedback} in the form of multi-view video streaming has been used in \cite{wei2021multi} where
multiple RGB-D cameras are merged to obtain complete coverage of the scene. This paper uses a VR headset to allow the user to view the point cloud from multiple angles with depth perception.
For bandwidth-constraint wireless applications, compression of either point clouds or RGB-D data is of importance, \cite{PCC_comparison} compares various approaches. \cite{PCSTREAM2021} presents applied point cloud compression and streaming to mobile devices. The Motion Picture Experts Group (MPEG), also involved in \textsl{H.264} is developing codecs for streaming point cloud data \cite{vpccgpcc}.
Utilizing edge computing to remove the burden of streaming and rendering the point cloud is a popular idea, \cite{Holo2022} describes a framework. Most codecs for depth data compression implement lossy compression. Jiffy Codec \cite{jiffycompression} provides a lossless approach to the compression of depth data coming from LIDAR or RGB-D sensors. It utilizes single instruction multiple data (SIMD) based integer compression.

From a networking perspective, \cite{XRteleop5G2021} gives a list of connectivity requirements for a teleoperation setup. \cite{fiveg_tactile} discusses 5G teleoperation-specific aspects. \cite{xiao2022} presents a comparison of networking methods for teleoperation, including 5G.

\section{DESIGN}
\label{sec:design}
\begin{figure*}
    \centering
    \includegraphics[width=150mm]{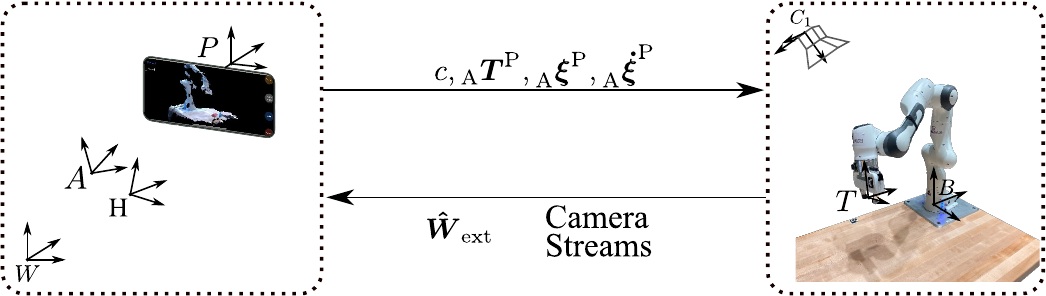}
    \caption{Leader and follower system and transmitted data streams. Also shown: Relevant frames.}
    \label{fig:system}
\end{figure*}

In the quest to implement a user-friendly system for mobile teleoperation of manipulators,
we observe three different dominating modalities: control of the manipulator position, 
visual feedback and feedback about contacts.
\paragraph*{Alteration of the manipulator pose} Conventional haptic devices for teleoperation provide
position-level continuous input in SE3. Any other approach such as velocity-level interfaces or discrete input (buttons) greatly reduces the intuitiveness. Especially when involving haptic feedback, it is crucial
to allow the user to react with reflexes to newly encountered contacts. 
In our approach, we want to replicate the user experience during kinesthetic teaching or hand guiding of a robot. Just that instead of grasping the tool, the user grasps the phone and moves the tool indirectly. A clutch button has to be pressed and held by the user to actually move the robot. 
As user observes the robot through the VR environment from a known position, the motions of the phone are applied to the robot in a way that
the manipulator shown in the VR environment moves in the same way. Hence, the input of the user is always relative to the current pose of the manipulator, but the mapping takes into account the absolute position of the user in the VR space.
\paragraph*{Control of manipulator position and integrated force control}
While optimal transparency can only be realized by small time constants in all subsystems, this imposes high requirements on the used devices and the network connecting them.
We replace the need for perfect tracking and force feedback with
a compliant control and simplified force feedback requiring a less complex leader device.
As there is no way to constrain the motion of the operator holding the phone, the operator
can push the robot into contact and create excessive contact forces when using a non-compliant controller.
In the presented approach, the compliant control approach implements an upper
limit to the contact forces. This is also relevant in case the pose estimated by the visual-inertial
fusion can occasionally contain outliers. As an added feature, a compliant control approach
allows the user to control the interaction force indirectly by pushing in the direction of a
contact constraint.
\paragraph*{Contact feedback} Information about contacts is necessary for effectively working within a teleoperation
environment. Again, our objective is to come as close as possible to 6D force feedback provided by
haptic interfaces using the available hardware on a mobile device. As only a single haptic
actuator is available in current devices, reproducing arbitrary static forces or force in multiple directions is impossible.
Hence, we designed a system which enables the user to sense transients related to the creation
of a first contact and additionally sense continuous forces through an amplitude modulation scheme.
All different force directions are mapped to a single haptics channel.
For high-frequency components of the forces created when establishing a new contact, we relay a transient haptic signal
which is scaled to the impulse the robot loses when making this contact. This removes the need for
high-frequency sensor-based contact force measurement and allows to rely on estimation of external forces.
For continuous contact forces, we relay a continuous haptic signal which is scaled with the magnitude of the
norm of the contact forces.
\paragraph*{Visual feedback} We opted for a multi-view live stream of the scene. The multi-view video stream is presented in a virtual reality environment which allows the user to reposition the view port by moving the phone. Motions of the user are mapped 1:1 into the virtual environment. Additionally, the design includes a clutch which temporarily disables the movement in the VR space. This allows the user to index through the space similar to indexing (activation of the clutch) for control or teleportation in other VR applications. We opted to capture both the scene and the robot with RGB-D cameras and aim to provide 360-degree coverage.  In our setup, we implement multi-view video by streaming separate colour and depth data from a number of \textsl{RGB-D} cameras to a phone where they are assembled into one point cloud. Extrinsic properties of
the cameras w.r.t. the robot are obtained beforehand.
\paragraph*{Connectivity} The objective is to propagate the real-time
streams shown in Fig.~\ref{fig:system} over a wireless network.  Those are the command (pose, velocity, acceleration, clutch status), and feedback consisting of the external forces and video streams. Each stream has its own properties in terms of bandwidth, acceptable latency and reliability.
Another design goal was to build a system which works in realistic network environments 
and reliably establishes a connection
in the presence of firewalls or other network obstructions. Teleoperation is a perfect example of a
peer-to-peer application and is directly impacted by those. \textsl{WebRTC}~\cite{webrtc} is a proven
design for video streaming with the option to add additional real-time data streams. It uses \textsl{UDP}
to encapsulate video streams and is able to work around said network obstructions using interactive connection establishment (ICE). Additionally,
it implements transport layer security encryption. Alternatives include \textsl{ZeroMQ} or \mbox{\textsl{ROS2/DDS}} but they provide only a subset of these features.

\section{CONTROL \& FEEDBACK}
\label{sec:control}
In this section, we describe the control scheme across the different components.
This comprises the leader and the follower side, refer to Fig.~\ref{fig:system}
for an overview. The following frames are defined:
$\text{W}$ is a world-fixed frame on the leader side,
$\text{P}$ is a frame attached to the mobile device,
$\text{B}$ is the robot base frame, and
$\text{T}$ is the robot tool frame. 
We proceed with the convention of denoting a transformation which transforms from frame $\text{Y}$ 
to frame $\text{X}$ as $\T{X}{Y} \in \text{SE}(3)$. The transformation can be defined using the translation
$\leftidx{_\text{X}}{\bs{p}}{^\text{Y}}$ and Rotation matrix $\leftidx{_\text{X}}{\bs{R}}{^\text{Y}}$, and
has the associated twist $\dT{X}{Y}$.
The meaning of the auxiliary frames $\text{A}$ and $\text{H}$ is described below, frame $\text{C}_1$ is the optical frame of the first camera. The Boolean clutch state is denoted as $c$.

\subsection{Visual feedback}
For visual feedback, the goal is to present the user with a freely navigable view of the scene, including the robot.
On the robot end, we use multiple RGB-D cameras to capture the scene. The camera positions $\T{B}{C,i}$ are known w.r.t. the robot's base.
We compress the colour image stream from each camera using the video codec \textsl{H.264}. For the depth stream compression
Jiffy compression \cite{jiffycompression} is used. After decompression, the known camera intrinsics are used to deproject the depth images to a point cloud. For indexing through the VR world, we introduce the frames $\text{A}$ and $\text{H}$. The position of these frames can be changed by the user by holding the "Hold view" button and moving the phone. During holding, the view port will not change. This allows the user to remain at one location and e.g. look at the scene from an opposite viewpoint without physically moving there.
%
\begin{align}
    \T{A}{P} &= 
        \begin{cases}
            \T{A}{H} &\text{hold view} \\
            \T{A}{W} \T{W}{P}  &\text{free}
        \end{cases}
    \\
    \T{A}{W} &=
        \begin{cases}
            \T{A}{H} \T{P}{W} &\text{hold view} \\
            \text{unchanged}  &\text{free}
        \end{cases}
    \\
    \T{A}{H} &=
        \begin{cases}
            \text{unchanged} &\text{hold view} \\
            \T{A}{P}  &\text{free}
        \end{cases}
    \\    
\end{align}
The point cloud of camera $i$ is shown in the VR environment at:
\begin{equation}
    \T{P}{C} = \T{P}{A} \T{B}{C,i}
\end{equation}
this shows that the frame $\text{A}$ in the VR space is equivalent to frame $\text{B}$ in the scene.
To obtain the $\T{W}{P}(t)$, a visual-inertial fusion is used. The description of
such is outside the scope of this paper. One possible approach is described in \cite{msckf}.
Initially, $\T{W}{P}=\bs{I}$ with the frame $\text{W}$  being gravity-aligned.

\subsection{Command}
The visual-inertial fusion additionally estimates the translation velocity $\bs{v}_\text{P}$. We can obtain the rotation velocity $\bs{\omega}_\text{P}$ and the translation acceleration $\bs{a}_\text{P}$ directly from sensor data. To move the manipulator, the user enables the clutch at time $t_\text{c}$ at the position $\T{A}{P}(t_\text{c})$. The Cartesian Impedance controller received the
following the desired pose:
\begin{align}
    \T{B}{T,d}(t) &= \T{B}{T}(0) \T{T}{D}(t)
\end{align}
with the offset $\T{T}{D}$ from the initial pose of the tool. When the clutch is engaged, $\T{T}{D}$ is updated using
\begin{align}
    \T{T}{D}(t) &= \T{T}{D}(t_\text{c}) 
    \begin{bmatrix}
        \bs{I} & - \leftidx{_\text{A}}{\boldsymbol{p}}{^\text{P}}(t_\text{c}) \\
        \bs{0} & 1
    \end{bmatrix}
    \T{A}{P}(t)
    \begin{bmatrix}
        \leftidx{_\text{P}}{\bs{R}}{^\text{A}}(t_\text{c}) & \bs{0} \\
        \bs{0} & 1
    \end{bmatrix}.
\end{align}
and left constant otherwise.

\subsection{Manipulator Control}
We assume the following rigid-body dynamics of the manipulator with the states $\bs{q}$ and $\bs{\dot{q}} \in \mathbb{R}^{N}$:
\begin{equation}
    \bs{M}(\bs{q})(\bs{\ddot{q}}) + \bs{C}(\bs{q},\bs{\dot{q}})\bs{\dot{q}} + \bs{g}(\bs{q}) = \bs{\tau} + \bs{J}^{\text{T}}(\bs{q}) \cdot \bs{W}_\text{ext}
\end{equation}
with the Mass matrix $\bs{M}$, Coriolis matrix $\bs{C}$, gravity terms $\bs{g}$, Jacobian $\bs{J}$ associated with the frame $\text{T}$ and external wrench $\bs{W}_\text{ext}$ acting on that frame. The actuator torques $\bs{\tau}$ are computed by a Cartesian impedance control approach with saturated force and torques:
\begin{align}
    \bs{\tau} &= \bs{J}^\text{T}
    \left[ 
        \bs{K}(\bs{e}) \cdot \bs{e} + \bs{D}(\bs{e}) \cdot \bs{\dot{e}} \right]  + \bs{\tau}_\text{ffwd} \\
     \bs{\tau}_\text{ffwd} &= \bs{C}(\bs{q},\bs{\dot{q}})\bs{\dot{q}} + \bs{M}(\bs{q}) \left[ \bs{J}^\text{T} \bs{\ddot{x}_\text{d}} + \bs{\dot{J}}^{\text{T}} \bs{\dot{x}_\text{d}}\right] \\
     \bs{K}(\bs{e}) &= \left[ \nicefrac{1}{\text{max}(0.1,\cosh(\bs{p}(\bs{e})))} \right]^2 \cdot \bs{K}_{\text{nom}} \\
     \bs{p}(\bs{e}) &= \bs{K}_{\text{nom}} \begin{bmatrix} \text{diag}(F_\text{max}) & \bs{0} \\ \bs{0} & \text{diag}(\tau_\text{max}) \end{bmatrix} \cdot \bs{e}
    \label{eqn:cartimp}
\end{align}
with the constant diagonal matrix nominal stiffness $\bs{K}_\text{nom}$, current
stiffness matrix $\bs{K}(\bs{e})$, the damping matrix $\bs{D}(\bs{e})$, and
feed-forward torques $\bs{\tau}_\text{ffwd}$. The error $\bs{e}$ is defined as
\begin{equation}
    \bs{e} =
    \begin{bmatrix}
        \leftidx{_\text{B}}{\bs{p}}{^\text{T,d}} - \leftidx{_\text{B}}{\bs{p}}{^\text{T}} \\
        \bs{r}_\text{err}( \leftidx{_\text{T}}{\bs{R}}{^{\text{B}}} \leftidx{_\text{B}}{\bs{R}}{^\text{T,d}}  )
    \end{bmatrix}
\end{equation}
with the position of the frame $\text{T}$ $\bs{p}_\text{T}$, orientation $\bs{R}_\text{T}$, and
angular error function $\bs{r}_\text{err}$.
$\bs{D}(\bs{e})$ derived using double-diagonalization based damping design~\cite{albu2003cartesian}
taking into account the current stiffness $\bs{K}(\bs{e})$.



\subsection{Contact Feedback}
We use a momentum-based external force observer~\cite{ObserverTRO} to obtain an estimate of $\bs{W}_\text{ext}$, $\bs{\hat{W}}_\text{ext}$:
\begin{align}
    \bs{\hat{W}}_\text{ext} &= \bs{J}^\text{T} \bs{\hat{\tau}}_\text{ext} \\
    \bs{\hat{\tau}}_\text{ext} &= \bs{K}_{\text{O}} \left[ \bs{M}(\bs{q})(\bs{\dot{q}}) - 
    \int \bs{\tau} - \bs{g}(\bs{q}) + \bs{C}^\text{T}(\bs{q},\bs{\dot{q}}) \bs{\dot{q}} + \bs{\hat{\tau}}_\text{ext} \text{d}s
    \right]
\end{align}
with $\bs{K}_\text{O}$ a diagonal matrix with observer gain.
For relaying the observed wrench $\bs{W}_\text{ext}$ to the user, we determine the haptic pattern to be played based on the contact state.
The contact state is determined with a hysteresis around the 2-norm of the forces $\bs{F}_\text{ext} \in \mathbb{R}^3$.

The event of creating a contact is relayed to the user by playing a haptic pattern with the intensity $i_\text{impulse}$ which
replicates the impulse the robot lost when creating the contact, expressed as the impulse lost in a certain time window $\Delta$ approximated by:
\begin{equation}
    i_\text{impulse} = i_\text{impulse,min} + \left|\bs{S} \bs{J} \bs{M} \bs{J}^T \bs{J} \left[\bs{\dot{q}}(t) - \bs{\dot{q}}(t - \Delta)\right]\right|_2
\end{equation}
with the minimum intensity $i_\text{impulse,min}$ and the selection matrix $\bs{S}$ which removes the impulse associated with the angular velocities.

After the contact has been created, the $\bs{F}_\text{ext}$ is mapped to a different cyclic haptic pattern for which
the intensity $i_\text{cyclic}$ scales with the norm of the external forces:
\begin{equation}
    i_\text{cyclic} = i_\text{cyclic,min} + \left[|\bs{F}_\text{ext}|_2 - F_\text{ext,thres}\right].
\end{equation}
with the minimum intensity $i_\text{cyclic}$ and the force threshold for contact activation $F_\text{ext,thres}$.
External torques are ignored in the current implementation.

\section{EXPERIMENTAL SETUP}
\label{sec:experimental}
This section will go over the different components of the setup, with notes about
implementation details and other insights.
\paragraph*{For the leader system} we selected an \textsl{Apple IPhone}~\cite{iphone11}
because of the low-latency and high-fidelity haptic actuator. For the Wi-Fi tests an Iphone 11 was used, while an Iphone 14 was used for the 5G tests. While the application can be realized as a web app using the \textsl{WebXR} API, there is no access to the haptic
actuator from within the browser sandbox. Hence, we implemented a native application. This additionally enables 
leveraging single-instruction-multiple-data (\textsl{SIMD}) parallelism for the compression of the depth data.

\paragraph*{For the follower system}
we implemented the control described in Sec.~\ref{sec:control} for a \textsl{Franka-Emika} collaborative robot~\cite{franka},
leveraging the provided \textsl{FCI}
interface to control the robot. Our application provides $\bs{\tau}$ to the robot.
The robot is controlled by a small form factor computer with an \textsl{Intel i7-8700T} CPU and a \textsl{NVidia Quadro P620} graphics card
which also handles all remaining computational tasks.
\paragraph*{Camera streams and compression}
Four \textsl{Intel Realsense D435} cameras are configured to $848\times480$ resolution at $30 \text{fps}$. The colour streams are compressed using
the \mbox{\textsl{NVENC}~\cite{NVENC}} hardware accelerated \mbox{\textsl{H.264}} video codec using the graphics card. On the phone,
the streams are decompressed using the \textsl{VideoToolbox} API, leveraging hardware acceleration. All colour stream compression is handled by \mbox{\textsl{ffmpeg}}.
The depth streams are compressed using the Jiffy codec~\cite{jiffycompression} which uses the \textsl{SIMD} CPU features for compression
and decompression. We reimplemented the Jiffy codec in C++ for portability reasons and changed the integer compression
library to \mbox{\textsl{TurboPFor}~\cite{TurboPFor}} to support ARM platforms.
We opted for $30 \text{fps}$ even though the cameras we are able to stream at $60 \text{fps}$ to avoid
saturating the phone CPU.

\paragraph*{Camera extrinsics} $\T{B}{C,i}$ is estimated first using hand-to-eye calibration. For cameras with significant overlap in point clouds, the extrinsic properties are refined using the iterative closest point algorithm. This allows for short setup time and flexible camera placement without a requirement for overlapping camera views, e.g. in typical extrinsic calibration approaches. 
\paragraph*{Impedance Controller} stiffness is set to $400 \nicefrac{\text{N}}{\text{m}}$ in translation, and $40 \nicefrac{\text{Nm}}{\text{rad}}$ in rotation using critical damping. The leader provides desired position $\T{B}{T}$,
desired velocity $\dT{B}{T}$ and desired acceleration $\ddT{B}{T}$ used in the controller~\eqref{eqn:cartimp}. The control approach is completed by a Nullspace controller regularizing the robot configuration.
We found that the compliant control approach utilizing feed-forward terms as much as possible also
reduces the sensitivity to transient errors in the estimated pose $\T{W}{P}$ which would otherwise
generate significant jumps or at least noise on the robot side. In the setup only
$\T{W}{P}$ is affected by those errors, as $\dT{B}{T}$ and $\ddT{B}{T}$ are directly derived from sensor data.
\paragraph*{External force and impulse observer} We use the external force observer data provided through the \textsl{FCI} interface. We found that in our tabletop setup, the estimate of external force can be inaccurate when the
robot is in motion. We are using a $10 \text{N}$ hysteresis activation threshold $F_\text{ext,thres}$ before starting to render contact forces using the haptic actuator. This avoids signalling non-existent contacts to the user.
\paragraph*{Haptic patterns} We have chosen a pattern for contact creation which feels like a sharp spike. For relaying a continuous contact force to the user, we chose a smoother continuously repeating pattern. We scale the patterns to cover a minimum and maximum static contact force between $10 N$ and $40 N$.
\paragraph*{Phone pose} We make use of the \textsl{ARKit} API provided by the phone manufacturer to estimate the phone pose $\T{W}{P}$.
See \cite{vio_benchmark} for a comparison of the pose estimation performance. The user motions are not scaled, i.e. the user motion is replicated by the robot 1:1. To obtain $\dT{W}{P}$ we directly use the output of the \textsl{CoreMotion} API which implements a filter which relies solely on data from proprioceptive sensors. For $\ddT{W}{P}$ the translation part is also provided by \textsl{CoreMotion} and we use a differentiating filter 
to obtain the angular acceleration. The update rates provided by the \textsl{ARKit} and \textsl{CoreMotion} API are 
$60 \text{Hz}$ and $100 \text{Hz}$ respectively. There are no additional filters implemented in the complete setup. Suitable environments are required for the phone pose estimation to work reliably. Covering or blinding the camera will lead to drift thus making the point cloud rendering drift out of the
user's view. Also, the command will be affected by this drift and this can lead to unintended motions.
\paragraph*{Network Connection}
For initial testing, we used a WiFi~4 access point which connects the phone to the robot. Alternatively, we connect to the Rogers \textsl{5G} non-standalone (NSA) test network available on campus using a \textsl{Quectel RM520N-GL} \textsl{5G} modem. To implement the \textsl{WebRTC} communication between leader and follower we use \textsl{libdatachannel}~\cite{libdatachannel}. Typically the used networks can exhibit a \textsl{stochastic packet loss} of below $1\%$ and have non-zero latency and jitter. Hence robustness of the application under these conditions is important. Both real-time data streams for the phone pose $\T{W}{P}$ and its derivatives and the estimated external force $\bs{W}_\text{ext}$ are fairly resilient to stochastic packet drops. However, video streams are typically more affected because for the computation of a frame, all fragments have to arrive. \textsl{H.264} is designed to be robust to a certain packet loss once an intra-coded frame is successfully received. The Jiffy Codec does not have these properties. Any dropped packet will make it impossible to allow computation of a frame until the next intra-coded frame arrives.
The application imprints different traffic classes in the differentiated services field of the IP packets, this allows network equipment to prioritize contact feedback streams over video streams. To reduce network congestion during high changes in the point cloud and thus increase traffic we impose an upper limit on video streams while prioritizing the other feedback traffic.
For the case of temporary loss of communication, i.e. drop of multiple packets in the control or force feedback
streams for $100\text{ms}$, the application is designed to disengage the clutch.

\section{PERFORMANCE EVALUATION}
\label{sec:performance}
\begin{figure}
    \includegraphics[width=\columnwidth,trim=0 100 200 0,clip]{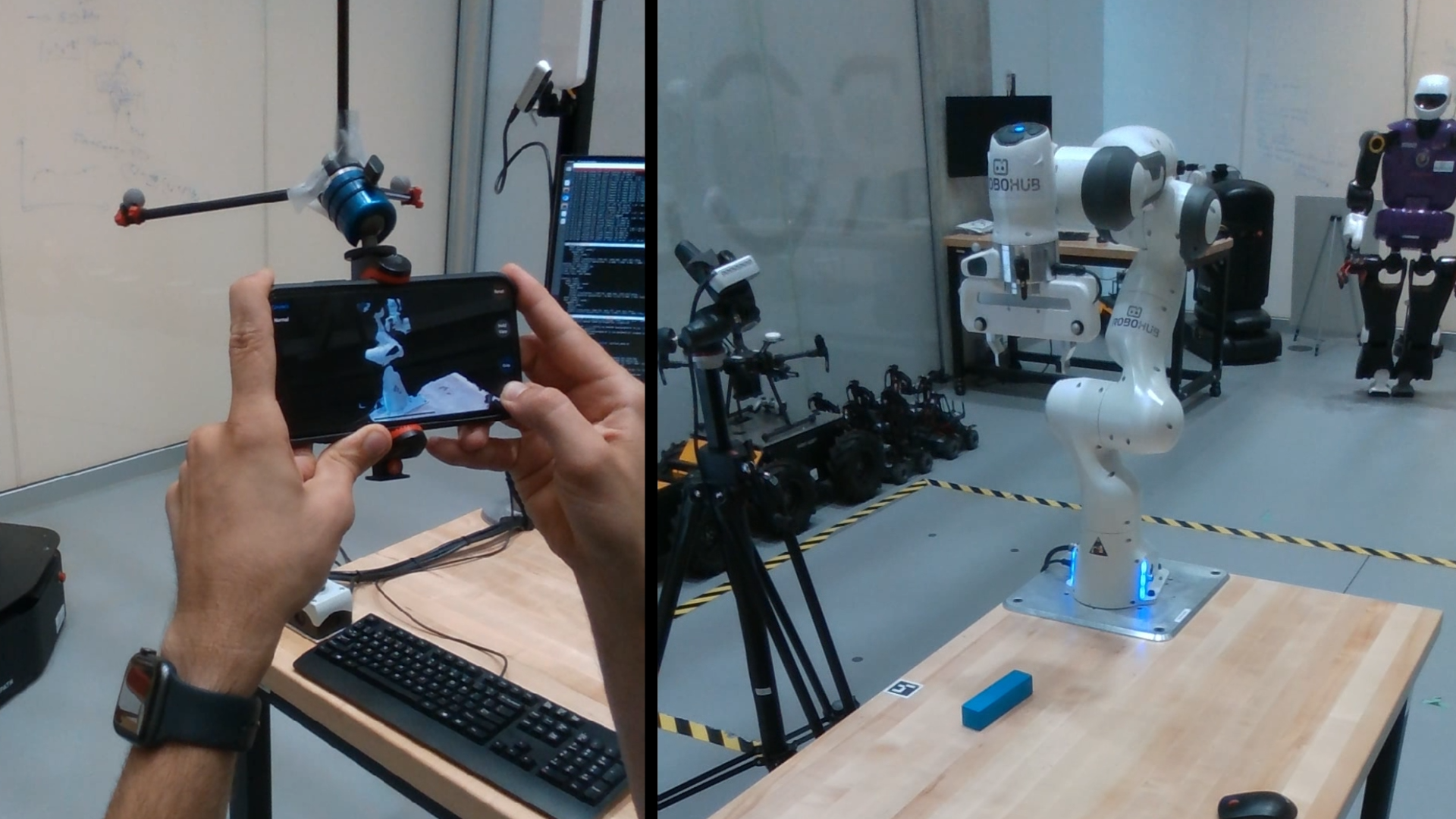}
    \caption{Evaluation setup: Manipulator and phone equipped with tracking markers.}
    \label{fig:evaluation_setup}
\end{figure}
\begin{figure}
    \centering
    \includegraphics{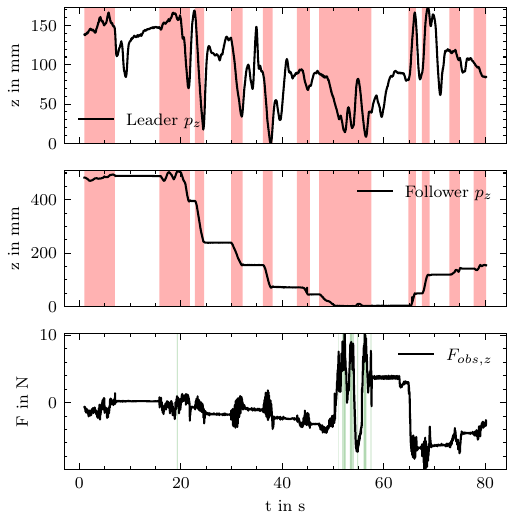}
    \caption{Example user session captured together with the accompanying video. The top plot shows the position of the leader device in
    $z$ direction, the middle plot the follower position. The bottom plot shows the estimated external forces. In the upper two plots the
    shaded areas are periods when the clutch is activated, in the lower plot the shaded areas denote an active contact.}
    \label{fig:tracking}
\end{figure}
For the performance evaluation, we created a local setup with near-optimal wireless connectivity.
We connected the phone to a wireless access point directly with minimal other traffic on this network.
In this setup, we observed an application-level round trip delay between leader and follower of 
a minimum of $5\text{ms}$, average of $20\text{ms}$, and $60\text{ms}$ maximum delay with minimal packet loss and reordering for Wi-Fi.
For 5G NSA we measured $40\text{ms}$, $66\text{ms}$, and $84\text{ms}$ respectively. For the final submission of the paper, this setup will completely move to 5G. The delay was measured using a separate \textsl{WebRTC} data channel, thus being affected by all user space, kernel, hardware and network aspects. As a qualitative result, we observed more consistent control over the manipulator position using 5G. We attribute this to the heavily utilized shared 5Ghz spectrum used by Wi-Fi at the test location.

For visual rendering, we observe an average delay created by compression, communication and decompression of
the colour images $50\text{ms}$ to $80\text{ms}$ in the Wi-Fi scenario.
A frame rate of $30\text{fps}$ is transmitted and rendered without a significant number of frames being
dropped.
To evaluate tracking performance without unnecessary artifacts we equipped both leader and
follower with markers for an external tracking system. This allows us to acquire both
the pose of the phone and the robot's tool frame with $1 kHz$ and no relative latency.

At steady state, we observe a traffic from leader to follower of $0.4 \nicefrac{\text{MBit}}{\text{s}}$ and 
$65 \nicefrac{\text{MBit}}{\text{s}}$ in the other direction. 
Bandwidth requirements of Jiffy codec depth stream heavily depend on depth noise and associated filter setup.
In general, the point cloud quality is the bottleneck for the user effectiveness in manipulation tasks
as adjacent objects are easily merged.
Visual feedback could significantly be improved with better sensor data.

In Fig.~\ref{fig:tracking} a typical user session is shown. The shown data was recorded during the session submitted as the accompanying video.
The user is making use of indexing to navigate the robot's workspace as
can be seen by the cycles the clutch status is changed. The motion of the user is contained in a relatively small comfort zone
(about $150mm$ in $z$). This is suitable for a user who does not use upper body motions as can be seen in the video. On the robot
side, the $z$ workspace is $500mm$. During the session, the user moves the robot into contact and receives haptic feedback about this.
In the bottom diagram of the figure, it can be seen that the user is taking note of the created contact and does not further
increase the contact force.

\begin{figure}
    \includegraphics{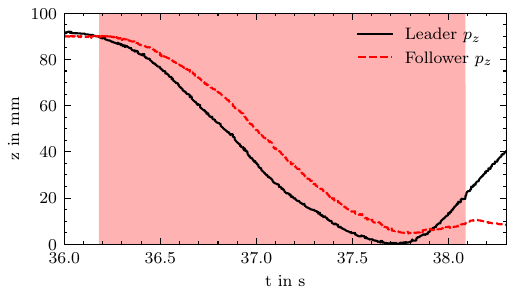}
    \caption{Tracking performance during a dynamic motion as recorded by the external tracking system.
            The user holds the clutch engaged during the area shaded in red.}
    \label{fig:tracking2}
\end{figure}
Fig.~\ref{fig:tracking2} shows tracking of leader and follower for a dynamic downwards motion. 
The data is recorded by the external tracking system which captures the pose of phone and tool, avoiding any misalignment of samples in time.



\section{CONCLUSION}
\label{sec:conclusion}
We described the design and implementation of a teleoperation scheme using a mobile phone as a leader device.
When the user moves the phone through space, the manipulator replicates the motions directly.
The phone pose is estimated using the built-in proprioceptive and
exteroceptive sensors
IMU and camera and transmitted over wireless networks using \textsl{WebRTC}. 
To realize an intuitive interface we implemented a compliant control scheme which uses the estimated pose
of the phone to control the manipulator pose. This scheme implements an upper limit to 
the static forces generated to eliminate potential damage to the environment or the robot.
For contact feedback, we map the
estimated external force on the manipulator to the haptic actuator on the phone. For visual feedback,
we stream multiple \textsl{RGB-D} cameras using hardware-accelerated compression schemes to the phone where
they are rendered into a point cloud. This enables the operator to freely navigate the scene and
overcomes the limitations of teleoperation interfaces with only single camera view and also
remedies the problem of missing depth perception on the phone display.

While some aspects of the mobile-phone-based teleoperation interface are clearly inferior to commercially available desk-mounted solutions for teleoperation of
manipulators, the focus here is to enable teleoperation with a readily available low-cost leader device. From any place, at any time, connecting to a robot which could also be mobile.
The strength of this approach lies in scenarios where teleoperation is used infrequently, e.g.
to resolve situations which are not correctly handled or not covered by an autonomous application. Alternatively,
this can be used as part of a shared-autonomy scheme for interactive manipulation.

To our surprise, we found that the visual feedback rather than the pose estimation is 
the bottleneck in this application.

\paragraph*{Future Work}
We observed significant development in the area of point cloud compression algorithms for real-time
multi-view video streaming. Together with better and more sensors, we hope to address the main bottleneck
of this approach.

To further improve immersion, we want to augment the displayed point cloud with an estimate of the contact point
locations and contact force direction indicators.

\addtolength{\textheight}{-8cm}   





\bibliographystyle{bibtex/IEEEtran}
\bibliography{bibtex/bibliography}

\begin{thebibliography}{10}
\providecommand{\url}[1]{#1}
\csname url@rmstyle\endcsname
\providecommand{\newblock}{\relax}
\providecommand{\bibinfo}[2]{#2}
\providecommand\BIBentrySTDinterwordspacing{\spaceskip=0pt\relax}
\providecommand\BIBentryALTinterwordstretchfactor{4}
\providecommand\BIBentryALTinterwordspacing{\spaceskip=\fontdimen2\font plus
\BIBentryALTinterwordstretchfactor\fontdimen3\font minus
  \fontdimen4\font\relax}
\providecommand\BIBforeignlanguage[2]{{%
\expandafter\ifx\csname l@#1\endcsname\relax
\typeout{** WARNING: IEEEtran.bst: No hyphenation pattern has been}%
\typeout{** loaded for the language `#1'. Using the pattern for}%
\typeout{** the default language instead.}%
\else
\language=\csname l@#1\endcsname
\fi
#2}}

\bibitem{wu2020development}
L.~Wu, R.~Alqasemi, and R.~Dubey, ``Development of smartphone-based human-robot
  interfaces for individuals with disabilities,'' \emph{Robotics and Automation
  Letters (RA-L)}, vol.~5, no.~4, pp. 5835--5841, 2020.

\bibitem{parga2013tele}
C.~Parga, X.~Li, and W.~Yu, ``{Tele-manipulation of robot arm with
  smartphone},'' in \emph{6th International Symposium on Resilient Control
  Systems (ISRCS)}.\hskip 1em plus 0.5em minus 0.4em\relax IEEE, 2013, pp.
  60--65.

\bibitem{teleopcomp2017}
D.~Kent, C.~Saldanha, and S.~Chernova, ``{A Comparison of Remote Robot
  Teleoperation Interfaces for General Object Manipulation},'' in
  \emph{International Conference on Human-Robot Interaction (HRI)}.\hskip 1em
  plus 0.5em minus 0.4em\relax ACM/IEEE, 2017, pp. 371--379.

\bibitem{handa2020dexpilot}
A.~Handa, K.~Van~Wyk, W.~Yang, J.~Liang, Y.-W. Chao, Q.~Wan, S.~Birchfield,
  N.~Ratliff, and D.~Fox, ``{Dexpilot: Vision-based teleoperation of dexterous
  robotic hand-arm system},'' in \emph{International Conference on Robotics and
  Automation (ICRA)}.\hskip 1em plus 0.5em minus 0.4em\relax IEEE, 2020, pp.
  9164--9170.

\bibitem{qin2023anyteleop}
Y.~Qin, W.~Yang, B.~Huang, K.~Van~Wyk, H.~Su, X.~Wang, Y.-W. Chao, and D.~Fox,
  ``{AnyTeleop: A General Vision-Based Dexterous Robot Arm-Hand Teleoperation
  System},'' in \emph{2023 Robotics: Science and Systems (RSS)}, 2023.

\bibitem{hetrick2020comparing}
R.~Hetrick, N.~Amerson, B.~Kim, E.~Rosen, E.~J. de~Visser, and E.~Phillips,
  ``Comparing virtual reality interfaces for the teleoperation of robots,'' in
  \emph{Systems and Information Engineering Design Symposium (SIEDS)}.\hskip
  1em plus 0.5em minus 0.4em\relax IEEE, 2020.

\bibitem{Impact2013}
J.~G. Wildenbeest, D.~A. Abbink, C.~J. Heemskerk, F.~C. van~der Helm, and
  H.~Boessenkool, ``{The Impact of Haptic Feedback Quality on the Performance
  of Teleoperated Assembly Tasks},'' \emph{Transactions on Haptics}, vol.~6,
  no.~2, pp. 242--252, 2013.

\bibitem{Cluttered2017}
J.~Bimbo, C.~Pacchierotti, M.~Aggravi, N.~Tsagarakis, and D.~Prattichizzo,
  ``{Teleoperation in cluttered environments using wearable haptic feedback},''
  in \emph{International Conference on Intelligent Robots and Systems
  (IROS)}.\hskip 1em plus 0.5em minus 0.4em\relax IEEE/RSJ, 2017, pp.
  3401--3408.

\bibitem{wei2021multi}
D.~Wei, B.~Huang, and Q.~Li, ``Multi-view merging for robot teleoperation with
  virtual reality,'' \emph{IEEE Robotics and Automation Letters (RA-L)},
  vol.~6, no.~4, pp. 8537--8544, 2021.

\bibitem{PCC_comparison}
F.~Pereira, A.~Dricot, J.~Ascenso, and C.~Brites, ``{Point cloud coding: A
  privileged view driven by a classification taxonomy},'' \emph{{Signal
  Processing: Image Communication}}, vol.~85, p. 115862, 2020.

\bibitem{PCSTREAM2021}
J.~Hu, A.~Shaikh, A.~Bahremand, and R.~LiKamWa, ``{Characterizing Real-Time
  Dense Point Cloud Capture and Streaming on Mobile Devices},'' in \emph{3rd
  ACM Workshop on Hot Topics in Video Analytics and Intelligent Edges}, ser.
  HotEdgeVideo '21.\hskip 1em plus 0.5em minus 0.4em\relax New York, NY, USA:
  Association for Computing Machinery, 2021.

\bibitem{vpccgpcc}
D.~Graziosi, O.~Nakagami, S.~Kuma, A.~Zaghetto, T.~Suzuki, and A.~Tabatabai,
  ``{An overview of ongoing point cloud compression standardization activities:
  video-based (V-PCC) and geometry-based (G-PCC)},'' \emph{APSIPA Transactions
  on Signal and Information Processing}, vol.~9, no.~1, 2020.

\bibitem{Holo2022}
P.~Qian, V.~S.~H. Huynh, N.~Wang, S.~Anmulwar, D.~Mi, and R.~R. Tafazolli,
  ``{Remote Production for Live Holographic Teleportation Applications in 5G
  Networks},'' \emph{IEEE Transactions on Broadcasting}, vol.~68, no.~2, pp.
  451--463, 2022.

\bibitem{jiffycompression}
{Jeff Ford and Jordan Ford}, ``{Lossless SIMD Compression of LiDAR Range and
  Attribute Scan Sequences},'' in \emph{{International Conference on Robotics
  and Automation (ICRA)}}.\hskip 1em plus 0.5em minus 0.4em\relax IEEE, 2023.

\bibitem{XRteleop5G2021}
F.~Hu, Y.~Deng, H.~Zhou, T.~H. Jung, C.-B. Chae, and A.~H. Aghvami, ``{A Vision
  of an XR-Aided Teleoperation System toward 5G/B5G},'' \emph{Communications
  Magazine}, vol.~59, no.~1, pp. 34--40, 2021.

\bibitem{fiveg_tactile}
Y.~Qiao, Q.~Zheng, Y.~Lin, Y.~Fang, Y.~Xu, and T.~Zhao, ``{Haptic
  Communication: Toward 5G Tactile Internet},'' in \emph{Cross Strait Radio
  Science \& Wireless Technology Conference (CSRSWTC)}, 2020, pp. 1--3.

\bibitem{xiao2022}
X.~Chen, L.~Johannsmeier, H.~Sadeghian, E.~Shahriari, M.~Danneberg, A.~Nicklas,
  F.~Wu, G.~Fettweis, and S.~Haddadin, ``{On the Communication Channel in
  Bilateral Teleoperation: An Experimental Study for Ethernet, WiFi, LTE and
  5G},'' in \emph{International Conference on Intelligent Robots and Systems
  (IROS)}.\hskip 1em plus 0.5em minus 0.4em\relax IEEE/RSJ, 2022, pp.
  7712--7719.

\bibitem{webrtc}
``{WebRTC: Real-Time Communication in Browsers},''
  \url{https://www.w3.org/TR/webrtc/}, 2023, [Online; accessed 2023-08-21].

\bibitem{msckf}
A.~I. Mourikis and S.~I. Roumeliotis, ``{A Multi-State Constraint Kalman Filter
  for Vision-aided Inertial Navigation},'' in \emph{International Conference on
  Robotics and Automation (ICRA)}, 2007, pp. 3565--3572.

\bibitem{albu2003cartesian}
A.~Albu-Schaffer, C.~Ott, U.~Frese, and G.~Hirzinger, ``{Cartesian impedance
  control of redundant robots: Recent results with the DLR light weight
  arms},'' in \emph{International Conference on Robotics and Automation
  (ICRA)}.\hskip 1em plus 0.5em minus 0.4em\relax IEEE, 2003.

\bibitem{ObserverTRO}
S.~Haddadin, A.~De~Luca, and A.~Albu-Schäffer, ``{Robot Collisions: A Survey
  on Detection, Isolation, and Identification},'' \emph{IEEE Transactions on
  Robotics (TRO)}, vol.~33, no.~6, pp. 1292--1312, 2017.

\bibitem{iphone11}
``{IPhone 11 - Technical Specifications},''
  \url{https://www.apple.com/by/iphone-11/specs/}, 2023, [Online; accessed
  2023-08-21].

\bibitem{franka}
``{Franka-Emika Robot Instruction Handbook},''
  \url{https://download.franka.de/documents/100010_Product\%20Manual\%20Franka\%20Emika\%20Robot_10.21_EN.pdf},
  2023, [Online; accessed 2023-08-21].

\bibitem{NVENC}
``{NVIDIA Video Codec SDK},''
  \url{https://developer.nvidia.com/video-codec-sdk}, 2023, [Online; accessed
  2023-08-21].

\bibitem{TurboPFor}
\textsl{powturbo}, ``{TurboPFor-Integer-Compression},''
  \url{https://github.com/powturbo/TurboPFor-Integer-Compression.git}, 2023.

\bibitem{vio_benchmark}
P.~Kim, J.~Kim, M.~Song, Y.~Lee, M.~Jung, and H.-G. Kim, ``{A Benchmark
  Comparison of Four Off-the-Shelf Proprietary Visual–Inertial Odometry
  Systems},'' \emph{Sensors}, 2022.

\bibitem{libdatachannel}
``{libdatachannel - C/C++ WebRTC network library},''
  \url{https://github.com/paullouisageneau/libdatachannel}, 2023, [Online;
  accessed 2023-08-21].

\end{thebibliography}

\end{document}